\documentclass[10pt,twocolumn,letterpaper]{article}

\usepackage[accsupp]{axessibility}
\usepackage{cvpr}              

\usepackage[T1]{fontenc}
\usepackage{multirow}
\usepackage{float}
\usepackage{graphicx}
\usepackage{tcolorbox}

\definecolor{cvprblue}{rgb}{0.21,0.49,0.74}
\usepackage[pagebackref,breaklinks,colorlinks,allcolors=cvprblue]{hyperref}

\def\paperID{35107} 
\def\confName{CVPR}
\def\confYear{2026}

\title{BridgeEQA: Virtual Embodied Agents for Real Bridge Inspections}

\author{
Subin Varghese, Joshua Gao, Asad Ur Rahman, Vedhus Hoskere\\
University of Houston\\
4226 MLK Blvd, Houston, TX 77204\\
{\tt\small \{srvargh2, jkgao, aurahman\}@cougarnet.uh.edu, vhoskere@central.uh.edu}
}

\newcommand{\numQAs}{2,200}
\newcommand{\numImages}{9,586}         
\newcommand{\avgImagesPerReport}{47.93}
\newcommand{\numReportsFiltered}{200}  
\newcommand{\numBridges}{200}          
\newcommand{\numTowns}{73}             
\newcommand{\imageMetricName}{Image Citation Relevance} 
\newcommand{\humanUpperLimitConditionRatingBaseline}{98\%}

\begin{document}
\twocolumn[{%
\renewcommand\twocolumn[1][]{#1}%
\maketitle
\begin{center}
    \centering
    \captionsetup{type=figure}
    \includegraphics[width=0.75\textwidth]{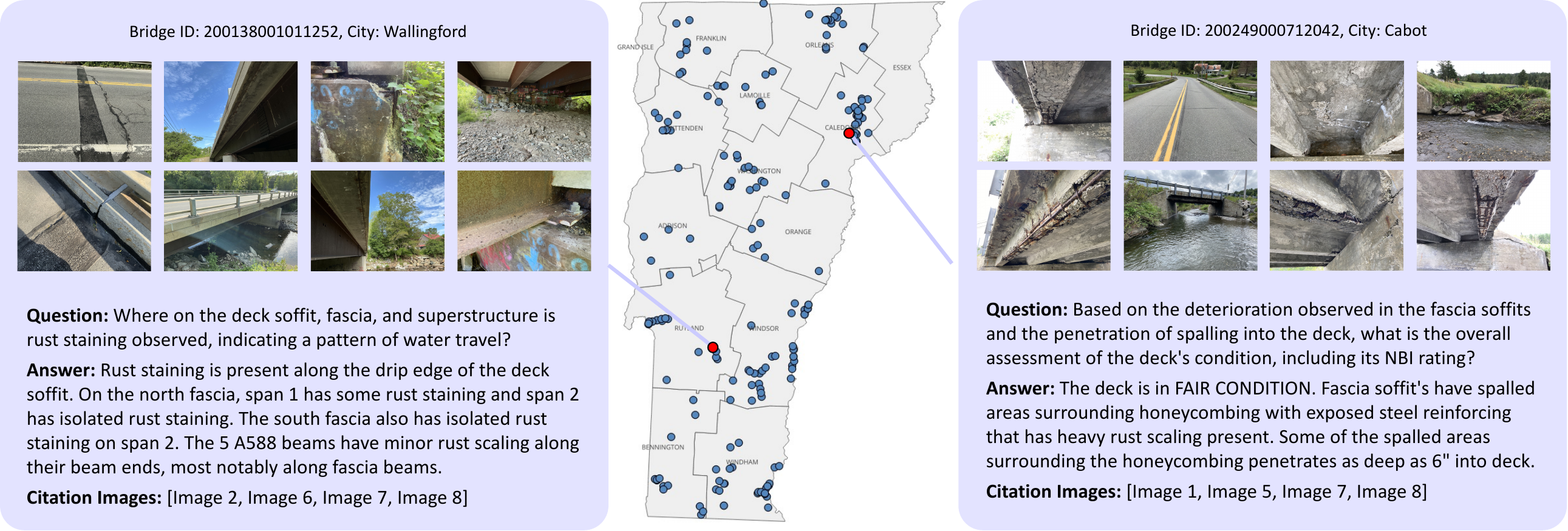}
    \captionof{figure}{\textbf{BridgeEQA: Open-Vocabulary Embodied Question Answering for bridge inspection.} Two example scenes from our benchmark showing questions that require synthesizing visual evidence across multiple egocentric images to assess bridges.}
\end{center}%
}]
\begin{abstract}
Deploying embodied agents that can answer questions about their surroundings in realistic real-world settings remains difficult, partly due to the scarcity of benchmarks for episodic memory Embodied Question Answering (EQA). Inspired by the challenges of infrastructure inspections, we propose Inspection EQA as a compelling problem class for advancing episodic memory EQA: it demands multi-scale reasoning and long-range spatial understanding, while offering standardized evaluation, professional inspection reports as grounding, and egocentric imagery. We introduce BridgeEQA, a benchmark of \numQAs{} open-vocabulary question-answer pairs (in the style of OpenEQA) grounded in professional inspection reports across \numBridges{} real-world bridge scenes with \avgImagesPerReport{} images on average per scene. We further propose a new EQA metric \imageMetricName{} to evaluate the ability of a model to cite relevant images. Evaluations of state-of-the-art vision-language models reveal substantial performance gaps. To address this, we propose  Embodied Memory Visual Reasoning (EMVR), which formulates the inspection EQA task as a Markov decision process. EMVR shows strong performance over the baselines. Code and dataset available at: \url{https://drags99.github.io/bridge-eqa/}
\end{abstract}
\section{Introduction}
\label{sec:intro}

When stacking a tower of blocks as a child, we learn not only to build upward but to probe structure: which elements are load-bearing, which are redundant, and how removing one piece will redistribute forces. After even a brief examination, we form a mental model of the tower's geometry and dependencies. Professional bridge inspectors exercise this form of spatial reasoning: moving through egocentric viewpoints, they synthesize visual evidence across components and time to assess structural condition with real consequences. This form of spatial reasoning strongly aligns with the task of Embodied Question Answering (EQA).

Recent embodied and spatial question answering benchmarks for vision-language models (VLMs) \cite{Yang2024ThinkingIS,das2018embodied,openeqa} tend to evaluate reasoning over small spatial extents and relatively simple queries, such as object counts or relative positions in constrained scenes. While these benchmarks are invaluable for measuring core capabilities, they under-represent challenges found in real-world deployments. These challenges include vast spatial extents, hierarchical organization from global overviews to fine-grained details, heterogeneous imaging conditions, and reconciling observations with domain-specific criteria. 

We propose infrastructure inspection, and bridge inspection in particular, as a compelling testbed for EQA in the style of Episodic Memory Embodied Question and Answering (EM-EQA) \cite{openeqa}, in which EQA is done over a pre-allocated set of images rather than active exploration. First, the domain naturally demands multi-scale reasoning, long-range spatial understanding, and complex semantic relationships between structural components that many times require multiple images to resolve. Second, a large volume of real-world data with expert annotations already exists in the form of professional inspection reports which include egocentric imagery and inspector notes regarding the structure. Third, standardized numerical ratings of components based on the National Bridge Inventory (NBI) scale \cite{fhwa1995coding} provide objective numerical values that can be used to evaluate agents' responses to directly compare to expert human inspectors. Finally, advancements in this domain have high potential for real-world impact as aging infrastructure requires regular, large-scale assessments that are labor-intensive and costly \cite{asce2021reportcard,fhwa1995coding}.

To this end, we introduce \textbf{BridgeEQA}, a benchmark of \numQAs{} open-vocabulary question-answer pairs, in the style of EM-EQA from \cite{openeqa} OpenEQA, with real imagery from bridge inspection reports across \numBridges{} bridge scenes, with an average of \avgImagesPerReport{} images per scene. Questions require aggregating visual evidence across multiple views and aligning responses with NBI condition ratings. Motivated by bridge inspection practice, where inspectors must justify numerical ratings with specific photographic evidence \cite{fhwa1995coding,nhi2006birm}, we evaluate condition rating accuracy and propose a new metric \textbf{\imageMetricName{}} that semantically evaluates the set of images a model cites to support its answer against a reference set of images. Finally, following existing open-vocabulary QA evaluation protocols \cite{openeqa,zheng2023judging,liu2024mmbench}, we also evaluate the open-vocabulary text response via LLM-as-a-judge \cite{zheng2023judging}. Together, these metrics holistically evaluate the model in terms of the alignment of open-vocabulary answers to ground truth answers, the relevance and faithfulness of the cited visual evidence, and agreement with expert human inspectors.

\begin{figure}
    \centering
    \includegraphics[width=0.45\textwidth]{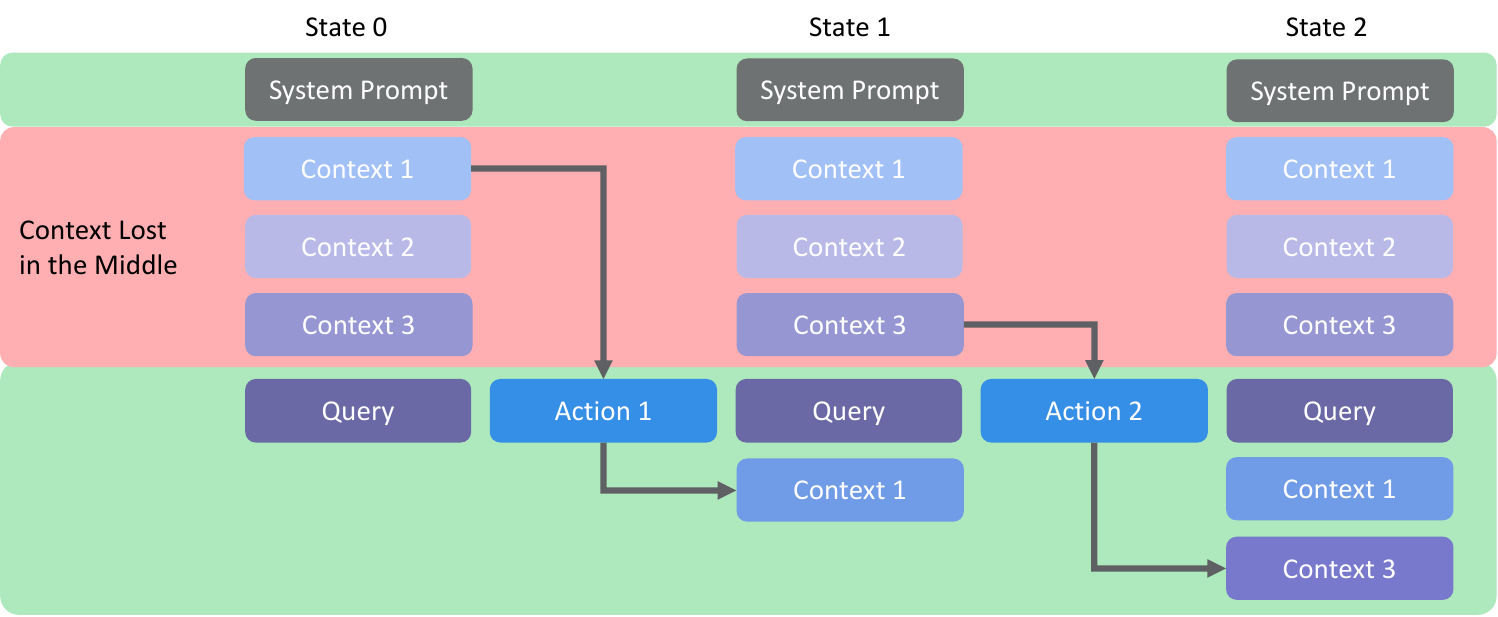}
    \caption{Illustration of how EMVR mitigates the ``lost in the middle'' problem. By navigating the scene graph and dynamically selecting relevant images, EMVR repositions critical visual evidence at the end of a VLM's context window, reducing mid-sequence information loss.}
    \label{fig:context_lost_in_the_middle}
\end{figure}

Evaluations using three state-of-the-art proprietary vision-language models, Gemini 2.5 Flash, Gemini 2.5 Flash-Lite, and Grok 4 Fast, with the strongest baseline method from OpenEQA for EM-EQA, Multi-Frame VLM \cite{openeqa}, revealed sizable performance gaps. Considering prior works have documented a positional bias in long-context LLMs toward the beginning or end of a sequence \cite{Liu2023LostITA,Bai2023LongBenchAB,Hsieh2024RULERWTB,Jiang2023LongLLMLinguaAAC,Kuratov2024BABILongTTD,An2024MakeYLE}, we theorized this may be the cause for poor performance. Therefore, we devised a reformulation of the Multi-Frame VLM approach for EM-EQA to be akin to an active Embodied agent in an Active EQA (A-EQA) setting. To do so, we direct the Embodied agent to dynamically retrieve context using a scene graph representation, in which images are nodes rather than objects, serving as an allocentric map. The Embodied agent must then make function calls to take actions such as to move to different nodes, analyze multiple images, analyze an image, and return a response in a Markov decision process (MDP). This dynamically allows the agent to select and promote mid-sequence information to the front of the context window, mitigating positional bias, Figure~\ref{fig:context_lost_in_the_middle}.
We call this method  Embodied Memory Visual Reasoning (EMVR) as it is akin to an agent reasoning over its memory. We find EMVR improves condition rating accuracy $\pm1$ by 9.34 percentage point, \imageMetricName{} by 20.2 percentage point, and Answer Correctness by 7.2 percentage point over Multi-Frame VLM using Grok 4 Fast.
In summary, this work makes the following contributions:
\begin{enumerate}
  \item \textbf{BridgeEQA}, a real-world EQA benchmark for infrastructure inspection with expert-grounded supervision, comprising \numQAs{} questions over \numImages{} images from \numBridges{} bridges across \numTowns{} towns, as an example of a new EQA problem class called \textbf{\textit{Inspection EQA}}.
  \item A new metric, \textbf{\imageMetricName{}}, for evaluating semantic similarity between agent-cited and reference images.
  \item \textbf{EMVR}, a novel EQA method that formulates QA as traversal over an image-based scene graph, improving condition rating accuracy by 13.6\%, visual evidence grounding by 29\%, and answer quality by 12.5\% over non-navigational baselines.
  \item Comprehensive baselines benchmarking contemporary VLMs and EQA methods on BridgeEQA.
\end{enumerate}

\section{Related Work}\label{sec:related_work}

\subsection{Bridge Inspection}

Bridge inspection is critical for structural safety and public welfare. According to the latest National Bridge Inventory (NBI) data, nearly 40~\% of U.S. bridges have reached or exceeded their 50-year design life, and over 10~\% are subject to load restrictions because structural deterioration limits heavy vehicle access~\cite{asce2021reportcard}. In the U.S., the National Bridge Inspection Standards (NBIS) mandate routine inspection of all public highway bridges longer than 20~ft at least every two years~\cite{fhwa1995coding}. During each routine inspection, certified inspectors evaluate components such as the deck, superstructure, and substructure and assign condition ratings on a standardized 0--9 scale to help prioritize maintenance and rehabilitation~\cite{fhwa1995coding,fhwa2011preservationguide}. Achieving consistent ratings is challenging because inspectors must correctly interpret rating guidelines and synthesize the effects of local defects across the structure into a component-level assessment~\cite{fhwa1995coding}. Although the guidelines are qualitative and rely heavily on inspector judgment, studies have found that expert inspectors typically agree within $\pm 1$ condition rating~\cite{agrawal2013consistency}. The Bridge Inspector's Reference Manual (BIRM) provides a step-by-step procedure for comparing field observations with rating criteria~\cite{nhi2006birm}. Recent advances in remote and autonomous inspection technologies~\cite{rakoczy2025technologies,ribeiro2025methodologies} and digital twin frameworks~\cite{hoskere2025unified} now offer promising platforms for implementing AI-assisted assessment systems. However, none provide an end-to-end autonomous solution for bridge inspections using only images of the bridge mimicking an inspector.

\subsection{Methods for Infrastructure Inspections}

Existing approaches can be broadly grouped into three capabilities: \emph{detection and classification}, \emph{visual question answering}, and \emph{report generation}. For detection, CLIP-based frameworks incorporating inspection knowledge, multi-view recognition models, and multi-class damage classifiers have been proposed to identify diverse defect types in bridge imagery~\cite{liao2024bridgeclip,varghese2024view,singh2025multiclass,varghese2025viewdelta, celik2025pixels}. Other work leverages few-shot CLIP for semantically guided UAV inspections, transformer-based multi-modal fusion for surface and subsurface damage segmentation, and instance segmentation of bridge point clouds~\cite{rahman2026bridgeelspect,malepati2025segmentation,rahman2025instance}. For VQA, bridge-specific vision-language pretraining using image-text pairs from inspection reports and multi-view VQA pipelines that integrate 3D reconstruction have been explored to support question answering and cause inference for observed damage~\cite{kunlamai2024vqa,yamane2024cause}. For report generation, image-to-text systems based on vision-language pretraining generate descriptive inspection narratives, while large-language-model-based frameworks produce structured maintenance plans from detected defects~\cite{wang2024automated,mohamed2025infragpt}. Beyond bridges, VQA benchmarks targeting post-disaster and remote sensing scenarios provide additional testbeds for assessing vision-language capabilities in infrastructure and environmental contexts~\cite{sarkar2023samvqa,rahnemoonfar2021floodnet,lobry2020rsvqa}. Despite these advances, no existing method evaluates the full reasoning chain that real inspection demands: navigating dozens of images spanning an entire structure, synthesizing cross-view evidence into component-level assessments, citing the supporting images, and aligning answers with codified inspection standards.

\subsection{Embodied Question Answering}

Embodied Question Answering (EQA) answers natural language questions about environments by reasoning over spatially distributed observations~\cite{das2018embodied, openeqa}. EQA encompasses two settings: episodic memory EQA (EM-EQA), where agents answer from pre-collected observations containing all required images, and \emph{active EQA} (A-EQA), where agents explore autonomously~\cite{openeqa}. The OpenEQA benchmark~\cite{openeqa} established the first open-vocabulary EQA dataset with 180 real-world scenes and 1600 QAs. Existing benchmarks focus predominantly on household environments with simple spatial layouts and queries (object counting, color identification), lacking the multi-scale structure, heterogeneous conditions with real imaging, and domain-specific criteria of professional inspection tasks \cite{das2018embodied,zhao2025cityeqa,dorbala2024house,azuma2022scanqa}.

Among EQA methods the Multi-Frame VLM~\cite{openeqa} has consistently shown to be the strongest baseline across open-vocabulary EQA benchmarks with varying domains~\cite{openeqa,zhao2025cityeqa,Ziliotto2024TANGOTE,li2025industryeqa}. As such we also use the Multi-Frame VLM method to establish a strong initial baseline. However, a weakness of this method is that all images are required as input to the VLM as context; as such, we theorize that this approach struggles with large image collections due to positional bias at long contexts where mid-sequence information is ``lost in the middle"~\cite{Liu2023LostITA, Bai2023LongBenchAB, Hsieh2024RULERWTB, Jiang2023LongLLMLinguaAAC, Kuratov2024BABILongTTD, An2024MakeYLE}. For inspection scenarios with potentially 100's of images in context, this bias would degrade answer quality and visual grounding drastically.

\subsection{Scene Graphs for Spatial Reasoning}

Scene graphs encode spatial and semantic relationships in environments, enabling symbolic reasoning for embodied agents. 3D scene graphs (3DSGs)~\cite{armeni2019scenegraph} organize scenes hierarchically to support navigation and manipulation, with recent frameworks~\cite{agia2022taskography, rana2023sayplan, gu2024conceptgraphs} using them to ground natural language instructions in spatial reasoning. Complementary 3D vision-language models~\cite{zhu2023vista, jia2025sceneverse, zhen2024vla} improve semantic grounding, though typically on point-cloud representations. In infrastructure inspection, recent work has enabled natural-language queries over point-cloud scenes~\cite{chen2024chat3d} and coordinated multi-agent drone inspection using 3D scene graphs~\cite{liu2025llminspection}. However, unlike general domains where robust object detectors and semantic segmentation models enable object-centric scene graphs, bridge inspection lacks foundation models capable of densely detecting all structural components (e.g., bearings, expansion joints, specific deterioration patterns). This limitation necessitates using images as graph nodes rather than detected objects. Motivated by these works and constraints, we use image-based scene graphs as allocentric maps for dynamic context retrieval.

\section{Methodology}\label{sec:methodology}

\subsection{Inspection EQA Problem Class}\label{subsec:inspection_eqa}

We define \emph{inspection EQA} as a general problem class: asset-centric, multi-view question answering in which an agent must synthesize visual evidence across multiple viewpoints of an inspected asset, align its answers to a standardized condition rubric, localize the supporting evidence, and achieve agreement with domain experts. While we instantiate this class on bridges, the formulation applies to any infrastructure asset with rubric-grounded, multi-view inspection data (e.g., dams, tunnels, pipelines). To make the problem class concrete and comparable across future domains, we propose a quantitative checklist. Any dataset that satisfies these properties at high rates instantiates an inspection EQA benchmark, enabling direct cross-domain comparison regardless of asset type. The checklist requires that all question-answer pairs depend on multiple views, that all answers are tied to a rating scale, that all pairs include reference image sets for evidence localization, and that citation relevance correlates strongly with human agreement.

\subsection{Scene Graph Formulation}\label{subsec:scene_graph}

We formalize bridge structures as navigable scene graphs constructed from inspection images. Crucially, this construction is purely visual and requires no GPS coordinates, geolocation metadata, or external spatial sensors. A VLM receives inspection images and outputs a structured JSON with minimal required fields (image description, central focus, edges), a design choice that promotes cross-domain generalizability. This structured representation enables the conversion of a set of images from an EM-EQA problem into an A-EQA problem allowing for systematic exploration by embodied agents.

\begin{figure}
  \centering
  \includegraphics[width=0.45\textwidth]{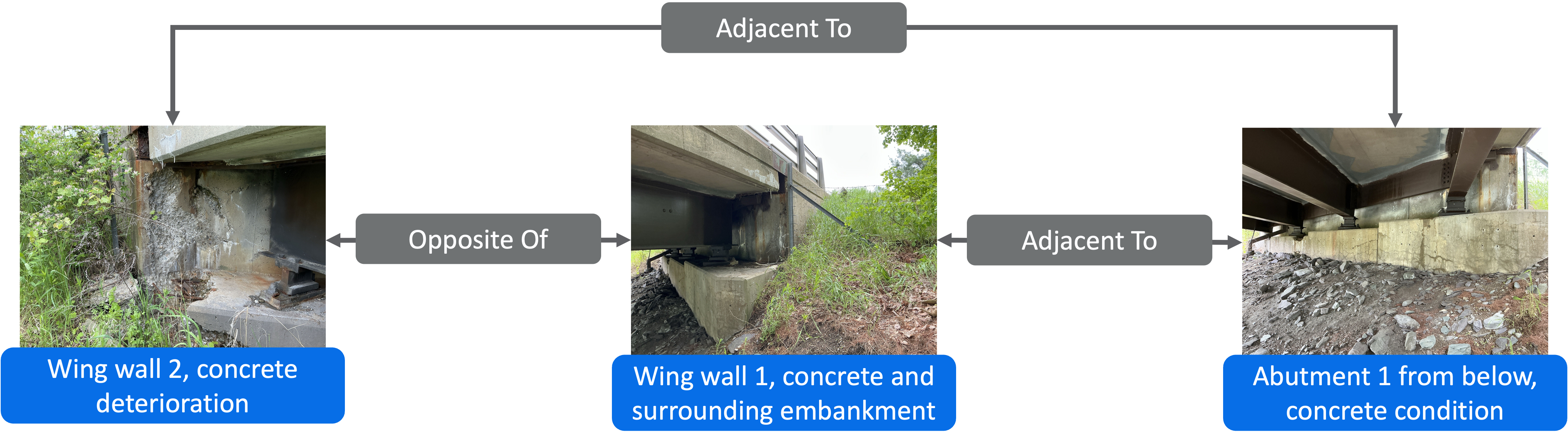}
  \caption{Illustrative example of scene graph structure for bridge inspection with image based nodes.}
  \label{fig:scene_graph}
\end{figure}

A scene graph $\mathcal{G} = (\mathcal{V}, \mathcal{E}, \mathcal{I})$ represents the physical bridge structure captured in inspection images, where:

\begin{itemize}
    \item $\mathcal{V}$ is a set of nodes, each representing a distinct viewpoint with an associated image
    \item $\mathcal{E} \subseteq \mathcal{V} \times \mathcal{V}$ is a set of directed edges representing spatial or semantic relationships between viewpoints
    \item $\mathcal{I}$ is the set of all images of the bridge, with a bijective mapping between nodes and images ($|\mathcal{V}| = |\mathcal{I}|$)
\end{itemize}

Each node $v \in \mathcal{V}$ encapsulates:

\begin{itemize}
    \item \textbf{Image name:} The associated photograph capturing the bridge structure
    \item \textbf{Central focus:} A semantic label describing the primary bridge component or viewpoint, using inspector terminology (e.g., ``Abutment 1 approach (South)'', ``Span 1 deck and superstructure'')
    \item \textbf{Image description:} Detailed visual observations of structural elements and conditions
    \item \textbf{Edge set:} Connections to semantically or spatially related nodes
\end{itemize}

Edges $(v_i, v_j) \in \mathcal{E}$ connect related viewpoints and include relationship descriptors (e.g., ``opposite approach'', ``supports span'', ``contains bearings''). This graph structure, Figure~\ref{fig:scene_graph}, transforms unordered image collections into spatially-organized representations of the bridge structure that embodied agents can navigate.

\textbf{Scene Graph Construction.} Scene graphs are constructed automatically using Gemini 2.5 Flash and fall back to Gemini 2.5 Pro when parsing errors are detected. The scene graph output as a JSON structure with a \texttt{nodes} array, where each node contains:

\begin{itemize}
    \item \textbf{image\_name:} Unique filename identifier for the image (e.g., \texttt{e23856c62ffb0.png})
    \item \textbf{central\_focus:} Concise semantic label for the primary bridge component or viewpoint (e.g., ``Superstructure steel girders and bearings at pier'')
    \item \textbf{image\_description:} Detailed visual observations of the bridge structure, including structural elements, defects, and contextual information (e.g., ``View of the superstructure showing steel open girders and cross-frames supported by ...'')
    \item \textbf{edges:} Array of directed edge objects, each containing:
    \begin{itemize}
        \item \texttt{connected\_to}: Target image filename
        \item \texttt{description\_of\_connection}: Natural language description of the semantic relationship
    \end{itemize}
\end{itemize}

The natural language edge descriptions capture the following relationship patterns:

\begin{itemize}
    \item \textbf{Hierarchical relationships:} Connecting overview and detail perspectives (e.g., ``is a detailed view of'', ``is an overview of a component detailed in'')
    \item \textbf{Structural relationships:} Physical support and load-bearing connections (e.g., ``supports'', ``is supported by'')
    \item \textbf{Spatial adjacency:} Neighboring components or locations (e.g., ``is adjacent to'', ``is an overview of the environment for'')
    \item \textbf{Condition similarity:} Viewpoints showing comparable defects or states (e.g., ``shows similar condition to'')
    \item \textbf{Component membership:} Part-whole relationships within larger assemblies (e.g., ``is a component of the deck shown in'')
\end{itemize}

We analyze the effect of node and edge count in the Supplementary Materials."

\subsection{ Embodied Memory Visual Reasoning}

\begin{figure}
    \centering
    \includegraphics[width=0.45\textwidth]{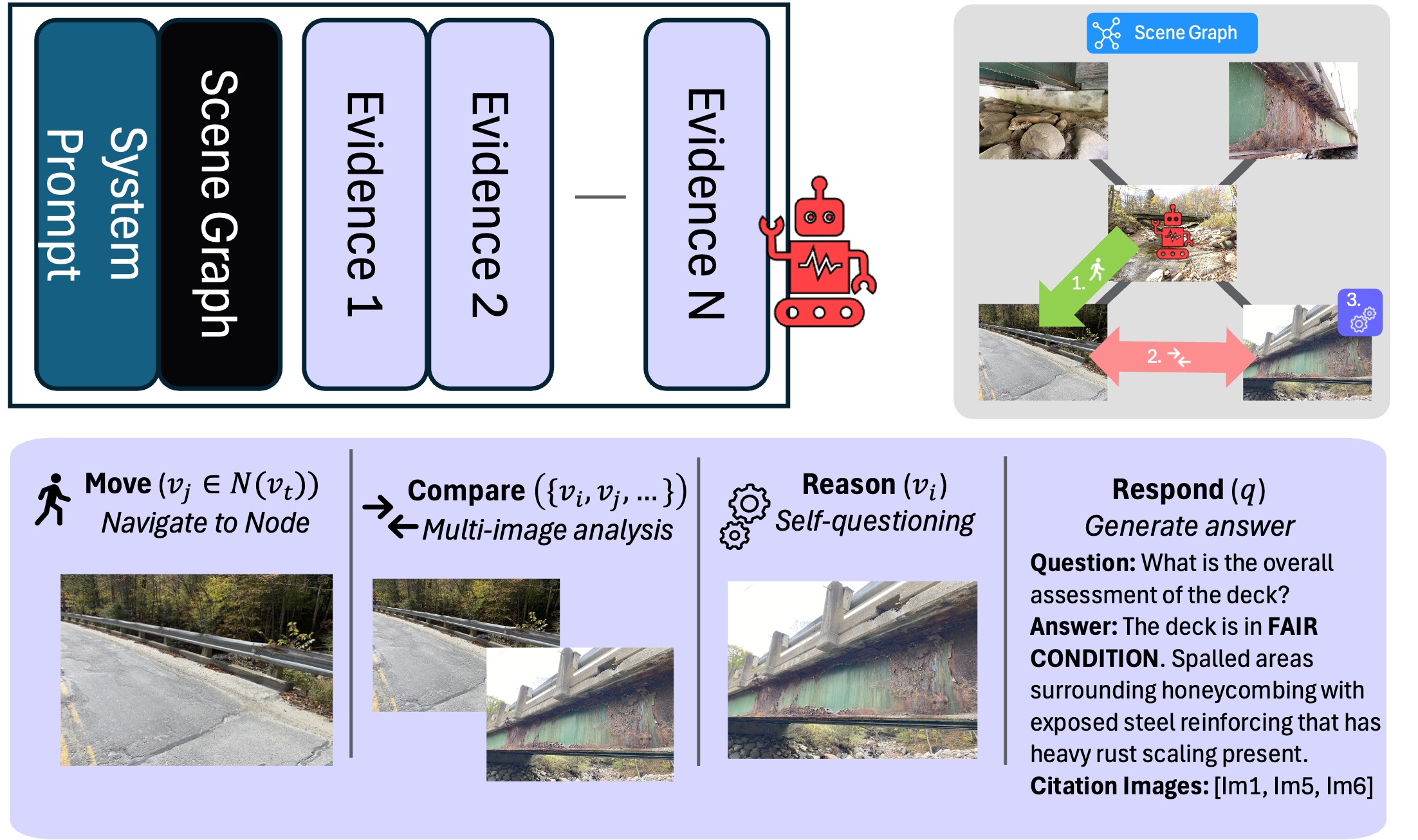}
    \caption{Overview of Embodied Memory Visual Reasoning in which an agent navigates a scene graph via an MDP, retrieving images dynamically to bring only relevant information into context.}
    \label{fig:framework}
\end{figure}

Embodied Memory Visual Reasoning frames the agent’s decision process as sequential navigation and selective recall, enabling it to retrieve and prioritize only the visual evidence needed to answer an inspection query. Figure~\ref{fig:framework} illustrates the complete EMVR framework: the scene graph $\mathcal{G}$ provides the structural context for selective image access. Unlike EM-EQA baselines that receive all images simultaneously and respond in a single pass, EMVR initializes with only the scene graph structure (nodes, edges, semantic labels) and then takes actions through an MDP to retrieve relevant visual context on demand.

We formulate the MDP as follows:

\begin{itemize}
    \item \textbf{State Space:} At time step $t$, the agent's state is $s_t = (v_t, h_t)$ where $v_t \in \mathcal{V}$ is the current node and $h_t$ represents the interaction history including previously viewed images and observations.

    \item \textbf{Observation Space:} The agent has access to the complete scene graph structure $\mathcal{G}$, including all node central focus labels, image descriptions, and edge relationships. At each time step, the agent observes its current node $v_t$ and can query neighboring nodes $\mathcal{N}(v_t) = \{v_j \mid (v_t, v_j) \in \mathcal{E}\}$.
    
    \item \textbf{Action Space:} The agent executes actions via function calls:
    \begin{itemize}
        \item $\textsc{Move}(v_j)$: Navigate to node $v_j \in \mathcal{N}(v_t)$, updating $v_{t+1} = v_j$
        \item $\textsc{Compare}(\{v_i, v_j, \ldots\})$: Load and analyze images from two or more nodes for comparative inspection, where $|\{v_i, v_j, \ldots\}| \geq 2$
        \item $\textsc{Reason}(v_i)$: Perform self-questioning on a single image at node $v_i$ to extract specific details
        \item $\textsc{Respond}(q)$: Generate an answer to the inspection query $q$ with cited image references and a condition rating, ending the trajectory.
    \end{itemize}
    
    \item \textbf{Policy:} A vision-language model implements policy $\pi(a_t \mid s_t, q)$ that selects function calls based on the current state and inspection query. The policy terminates upon executing $\textsc{Respond}$.
\end{itemize}


\subsection{Condition Rating Accuracy}

The NBI condition rating scale \cite{fhwa1995coding} ranges from \textbf{0 (Failed)} to \textbf{9 (Excellent)}, with each integer representing a distinct condition category based on observable structural characteristics. We report \emph{exact match} accuracy and \emph{within $\pm$1} accuracy. Exact matches between condition ratings for human inspectors are noisy, but there is a high agreement at $\pm$1 \cite{agrawal2013consistency,fhwa2001reliability} making it a more robust measure.

\subsection{\imageMetricName{}}
\begin{figure}
    \centering
    \includegraphics[width=0.8\linewidth]{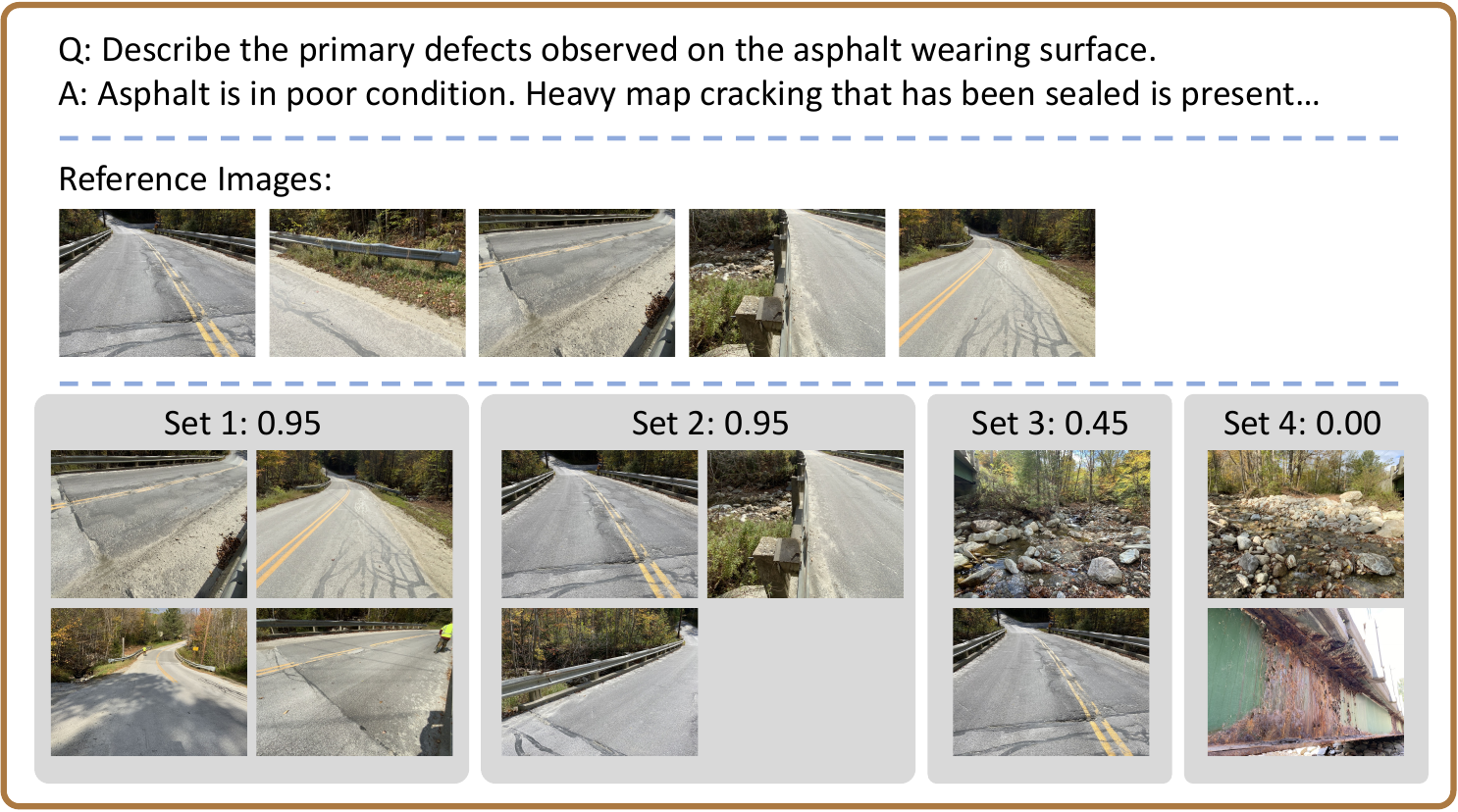}
    \caption{Example \imageMetricName{} scores for varying image citation sets in which multiple sets can be valid.}
    \label{fig:image_citation_relevance}
\end{figure}
Bridge inspectors justify condition ratings with photographic evidence. Similarly, \imageMetricName{} evaluates whether agents cite appropriate supporting images by semantically comparing agent selections against a reference set, Figure~\ref{fig:image_citation_relevance}. To achieve this, we employ a VLM-as-a-judge approach with Gemini 2.5 Flash, chosen for it's cost effectiveness and alignment with human preferences. The judge receives the question, the ground truth answer, reference images $\mathcal{R}$ (as examples, not definitive ground truth), and agent-selected images $\mathcal{R}_{\text{agent}}$, then scores on a 0.0–1.0 scale while penalizing over-selection in the event that an agent cites more than 5 times the number of images in the reference set. On average, all EQA methods chose fewer than 6 images; as such, they were never penalized sharply. The final \imageMetricName{} score averages judge ratings across all evaluation questions. We validate this metric for human alignment using three annotators, showing a Spearman correlation of 0.817 between the averaged human annotations and the \imageMetricName{} score. Additional details on the evaluation of this metric against human alignment are provided in the Supplementary Materials.

\textbf{Reference Image Citations} During dataset construction (Section~\ref{subsec:dataset_construction}), each question-answer pair is annotated with a set of reference images $\mathcal{R} = \{i_1, i_2, \ldots, i_k\}$ that provide visual evidence for the ground truth answer. These references are extracted from the original PDF inspection reports, where inspectors explicitly link textual condition descriptions to specific photographs. This annotation ensures that reference images represent inspector-validated visual evidence.

\textbf{Agent Image Citations} When answering an inspection query $q$, the agent generates a structured response that includes both a textual answer and an explicit list of supporting reference images $\mathcal{R}_{\text{agent}} = \{i'_1, i'_2, \ldots, i'_m\}$. This structured output format requires agents to explicitly cite which images provide visual evidence for their condition assessment, mirroring the documentation requirements of professional inspection reports.

\section{Dataset}\label{sec:dataset}

The BridgeEQA dataset comprises \numReportsFiltered{} bridge inspection reports from the Vermont Agency of Transportation (VTrans), spanning \numTowns{} Vermont towns with \numImages{} images (avg. \avgImagesPerReport{} per report) and \numQAs{} question-answer pairs annotated with NBI condition ratings. We split our dataset into a train and test set of 1,100 QA pairs each.

\subsection{Dataset Construction}\label{subsec:dataset_construction}
\begin{figure}
  \centering
  \includegraphics[width=0.45\textwidth]{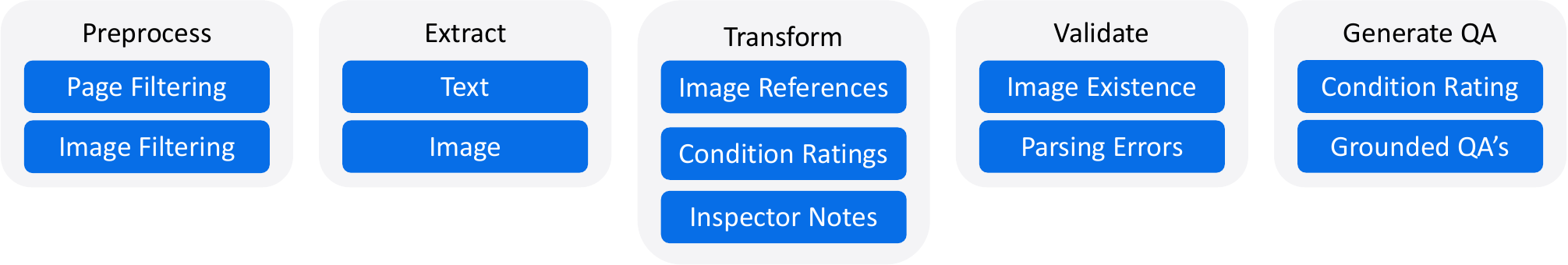}
    \caption{Pipeline for constructing the BridgeEQA dataset from Vermont Agency of Transportation (VTrans) inspection reports.}
  \label{fig:data_creation}
\end{figure}

We construct our dataset from unstructured PDF bridge inspection reports in the Vermont Agency of Transportation (VTrans) public database, where each report documents a single bridge with condition ratings, inspector notes, and photographs; the overall pipeline is summarized in Figure~\ref{fig:data_creation}. After applying report-level, page-level, and image-level quality filters, including a minimum threshold of 20 images per report and the removal of low-information pages and thumbnails, we randomly sample \numReportsFiltered{} reports and extract both textual and visual content that meets these criteria, yielding \numImages{} images and an average of \avgImagesPerReport{} images per report.

In the transformation and validation stages, we use Gemini 2.5 Flash and Gemini 2.5 Pro as zero-shot parsers to structure text and images, map image references to inspector notes, and extract NBI condition ratings while preserving inspector rationale, with Gemini 2.5 Pro serving as a fallback when parsing errors or hallucinations are detected. We then validate the structured data with automated checks and human review, generate grounded QA pairs with image references and condition labels, and evaluate QA quality using RAGAs Faithfulness \& Answer Relevancy \cite{es-etal-2024-ragas}, RAGalyst Answerability \cite{gao2025ragalyst}, and an LLM-as-a-Judge Inspector Relevancy score. We achieve a Faithfulness of 0.997, an Answer Relevancy of 0.997, an Answerability of 0.996, and an Inspector Relevancy of 0.980. Additional implementation and validation details are provided in the Supplementary Materials.

\subsection{Geographic and Structural Coverage}

\begin{figure}
  \centering
  \includegraphics[width=0.4\textwidth]{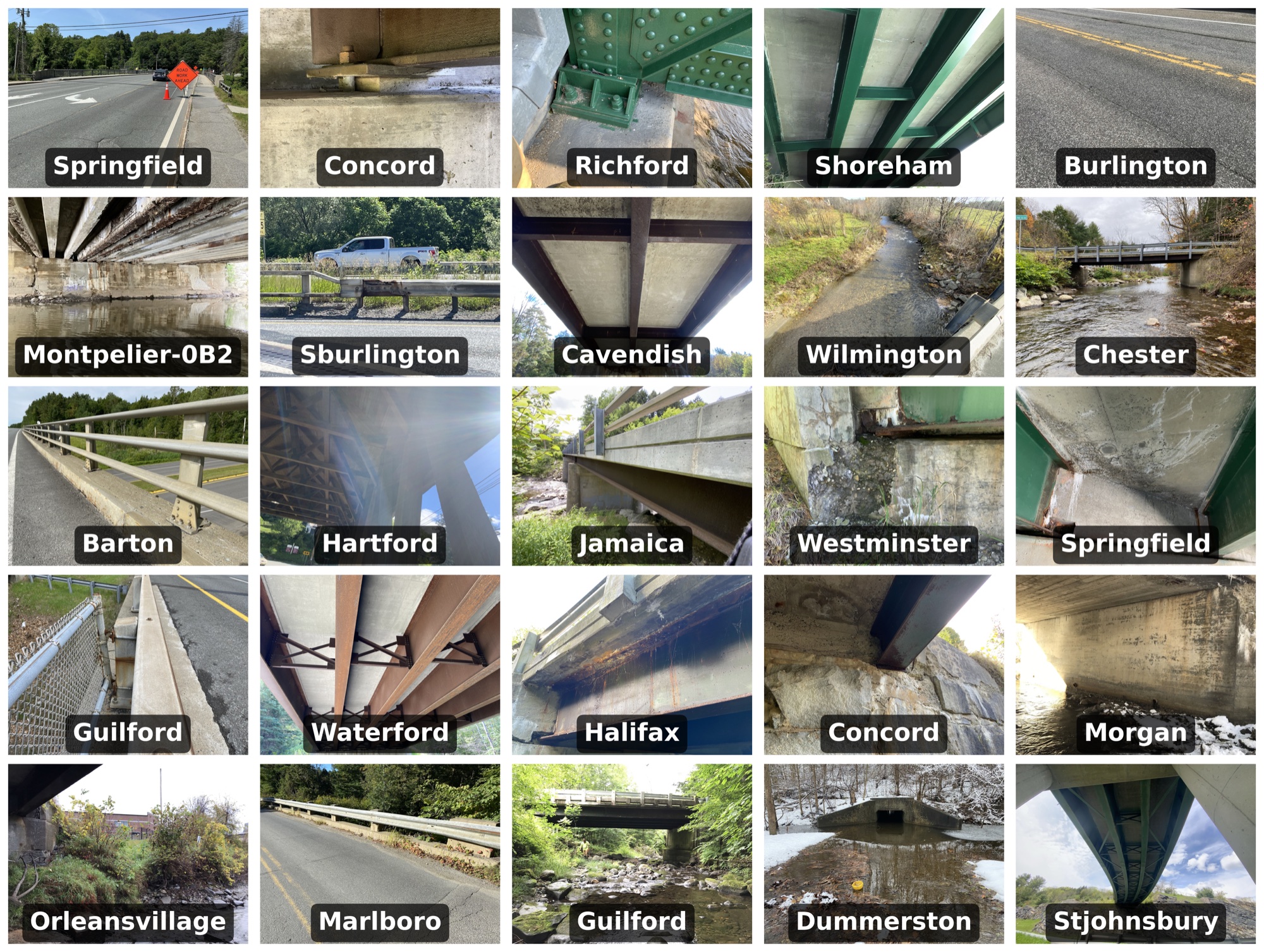}
  \caption{Representative sample images from BridgeEQA across Vermont bridges, demonstrating diverse bridge types (beam, truss, arch), construction materials, environmental conditions, and imaging perspectives.}
  \label{fig:bridge_diversity}
\end{figure}

The \numReportsFiltered{} inspection reports provide diverse coverage across bridge types (beam, truss, arch), construction materials (concrete, steel, timber, composite), environmental contexts (rural to urban, varying climatic exposure), and traffic conditions (low-volume rural routes to state highways), as shown in Figure~\ref{fig:bridge_diversity}. This diversity ensures models must generalize across varied contexts rather than overfit to specific bridge archetypes or environmental conditions.

\subsection{Condition Rating Distribution}

\begin{figure}
  \centering
  \includegraphics[trim={0mm 0mm 0mm 10mm}, clip, width=0.45\textwidth]{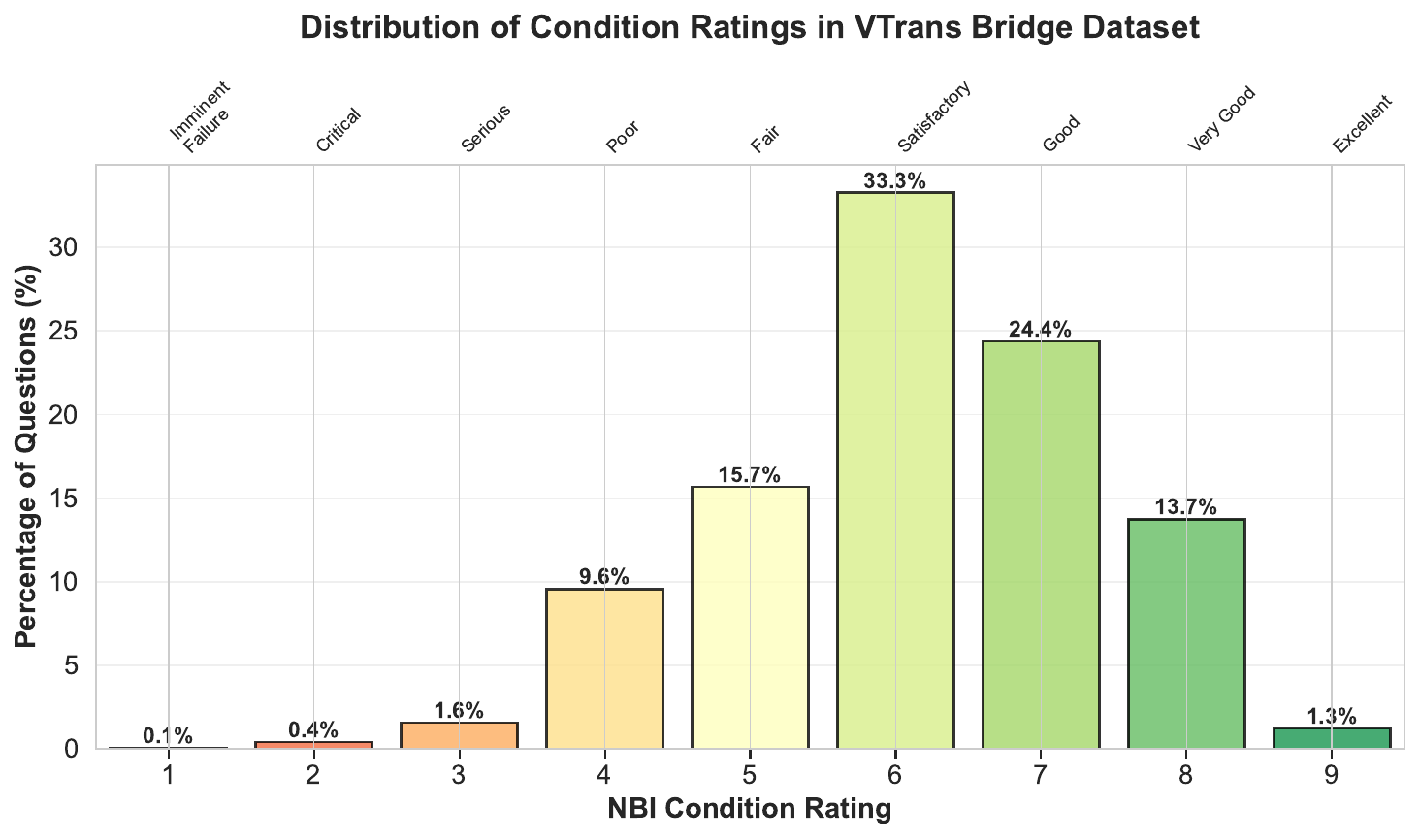}
  \caption{Distribution of NBI condition ratings for bridge components in the BridgeEQA dataset.}
  \label{fig:condition_rating_dist}
\end{figure}

The BridgeEQA dataset focuses on component-level condition assessments of bridge elements (decks, superstructures, substructures, abutments, wingwalls) using the NBI scale from 0 to 9, where higher ratings indicate better condition~\cite{fhwa1995coding}.

As shown in Figure~\ref{fig:condition_rating_dist}, the distribution is centered around ratings 5-7 (Fair to Good), with rating 6 (Satisfactory) being most common at 397 questions. This reflects the typical condition profile of aging infrastructure, where most components show minor deterioration but remain structurally sound. The dataset includes examples across the full rating spectrum (severely deteriorated components at ratings 1-4, excellent condition at ratings 8-9), enabling comprehensive evaluation of vision-language models' ability to assess diverse infrastructure conditions.

\subsection{Question Type Categorization}

\begin{figure}
  \centering
  \includegraphics[trim={10mm 50mm 10mm 50mm}, clip, width=0.3\textwidth]{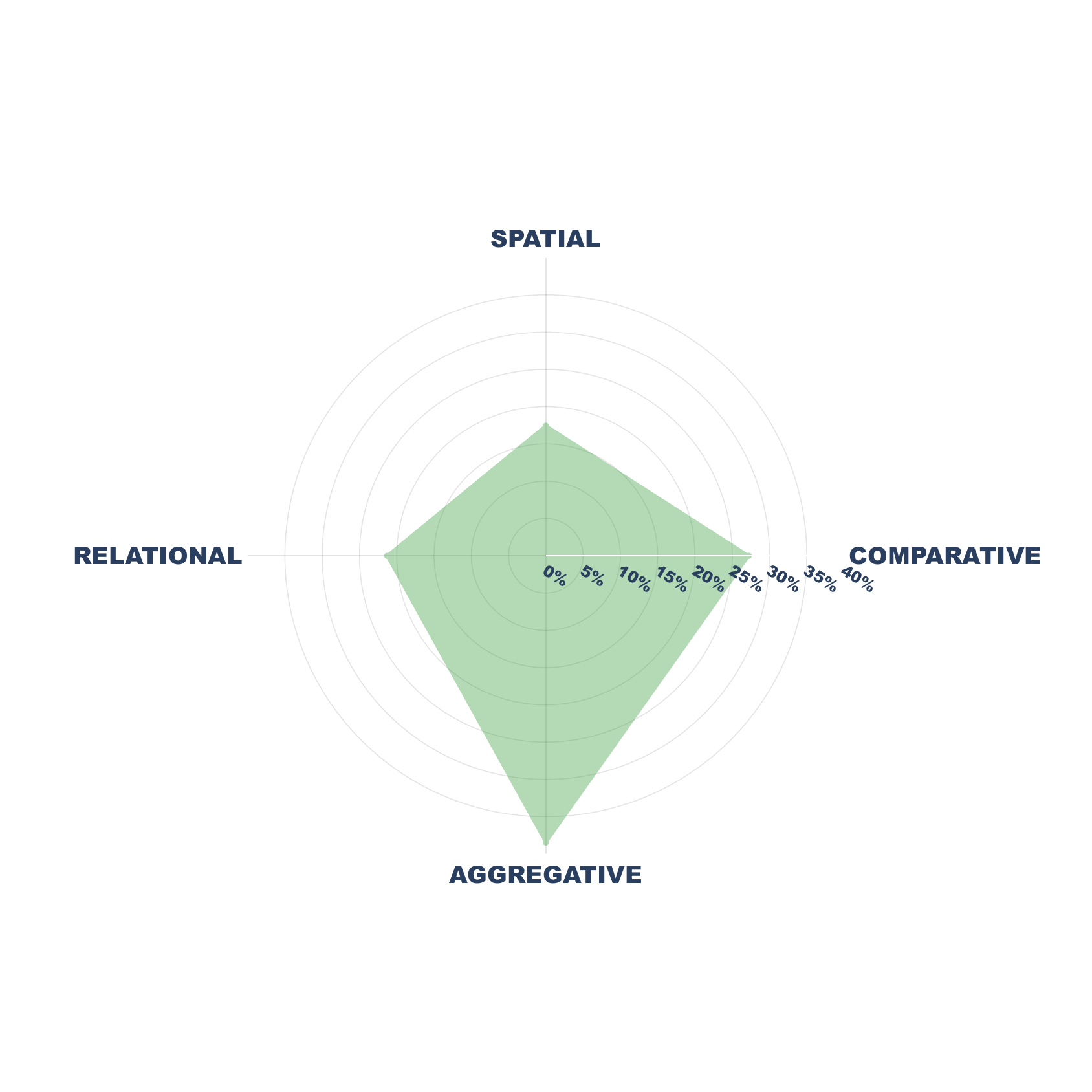}
  \caption{Distribution of question types in the BridgeEQA dataset (based on a random sample of 300 QA pairs). Each question can have multiple types simultaneously (e.g., both Comparative and Spatial), so percentages represent the proportion of questions containing each type and do not sum to 100\%.}
  \label{fig:question_type_radar}
\end{figure}

To ensure question diversity we randomly sample 300 QA pairs and categorize the types of questions in BridgeEQA into one or more of the following types:

\begin{enumerate}
  \item \textbf{Comparative:} Side-by-side comparison of structural elements (e.g., ``Compare cracking severity on upstream versus downstream pier faces'').

  \item \textbf{Spatial:} Location and distribution of deterioration patterns (e.g., ``Where is spalling most concentrated on the deck surface?'').

  \item \textbf{Relational:} Cause-effect reasoning about deterioration mechanisms (e.g., ``What caused the corrosion on the beam ends near the joint?'').

  \item \textbf{Aggregative:} Reasoning across multiple defect observations to form an overall condition assessment (e.g., ``Considering the spalling, cracking, and exposed rebar, what is the overall deck condition rating?'').
\end{enumerate}

As shown in Figure~\ref{fig:question_type_radar}, aggregative reasoning (38.5\%) and comparative analysis (27.2\%) are most prevalent, with relational reasoning (21.3\%) and spatial analysis (17.5\%) also represented. Questions frequently combine multiple types.

\section{Experiments}\label{sec:experiments}

\begin{figure*}
    \centering
    \includegraphics[width=0.8\linewidth]{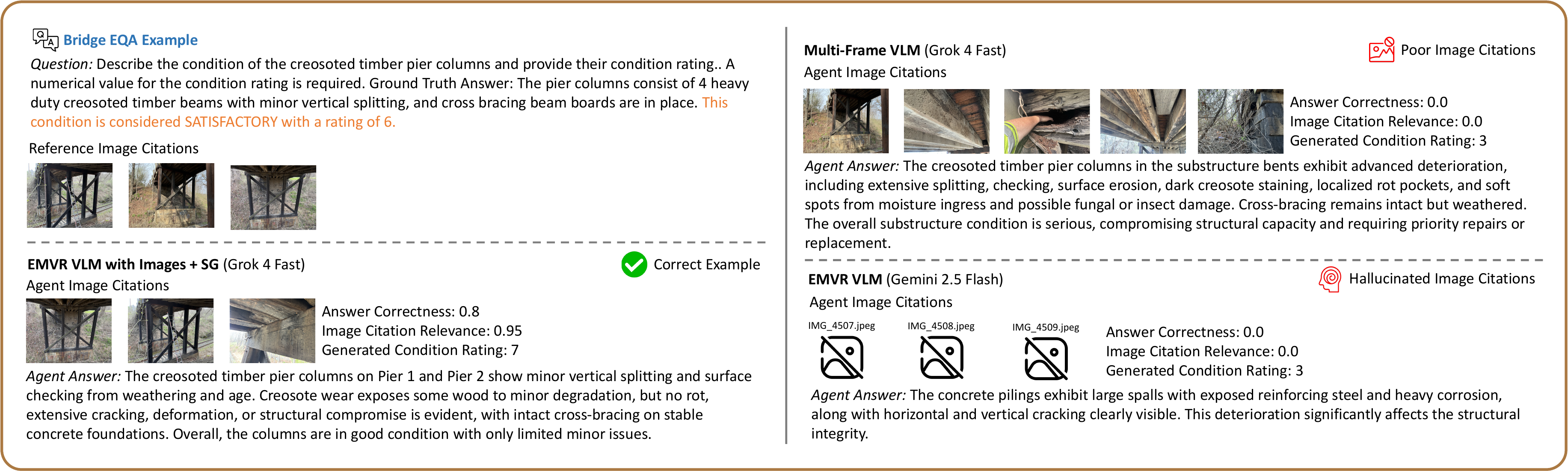}
    \caption{We illustrate a correct example alongside two common failure cases, poor image citation and hallucinated image citations. These suggest that low quality image citations can serve as a proxy for detecting hallucinations or poor answer generations.}
    \label{fig:qualitative}
\end{figure*}
\subsection{Experimental Setup}\label{subsec:exp_setup}

We evaluate five EQA methods following an experimental protocol aligned with prior open-vocabulary EQA work~\cite{openeqa,zhao2025cityeqa,Ziliotto2024TANGOTE,li2025industryeqa}. As strong baselines, we include \textbf{Multi-Frame VLM}~\cite{openeqa} and \textbf{Socratic LLM w/ SG}~\cite{openeqa,zeng2022socratic}, which have demonstrated consistently strong performance on existing open-vocabulary EQA benchmarks. We further augment Multi-Frame VLM with scene graph context, denoted \textbf{Multi-Frame VLM w/ SG}~\cite{openeqa}. In addition, we compare against EMVR with scene graphs only as initial context, \textbf{EMVR VLM w/ SG Only}, and with both images and scene graphs, \textbf{EMVR VLM w/ Images + SG}. To assess generalization across VLMs, we evaluate all methods with \textbf{Gemini 2.5 Flash Lite}, \textbf{Gemini 2.5 Flash}, and \textbf{Grok 4 Fast} on our test set of 1,100 QA pairs. To ensure fair comparisons, all methods were given the same prompt with context related to bridge inspections.

\subsection{Quantitative Results}\label{subsec:quantitative_results}

\begin{figure}
    \centering
    \includegraphics[width=0.65\linewidth]{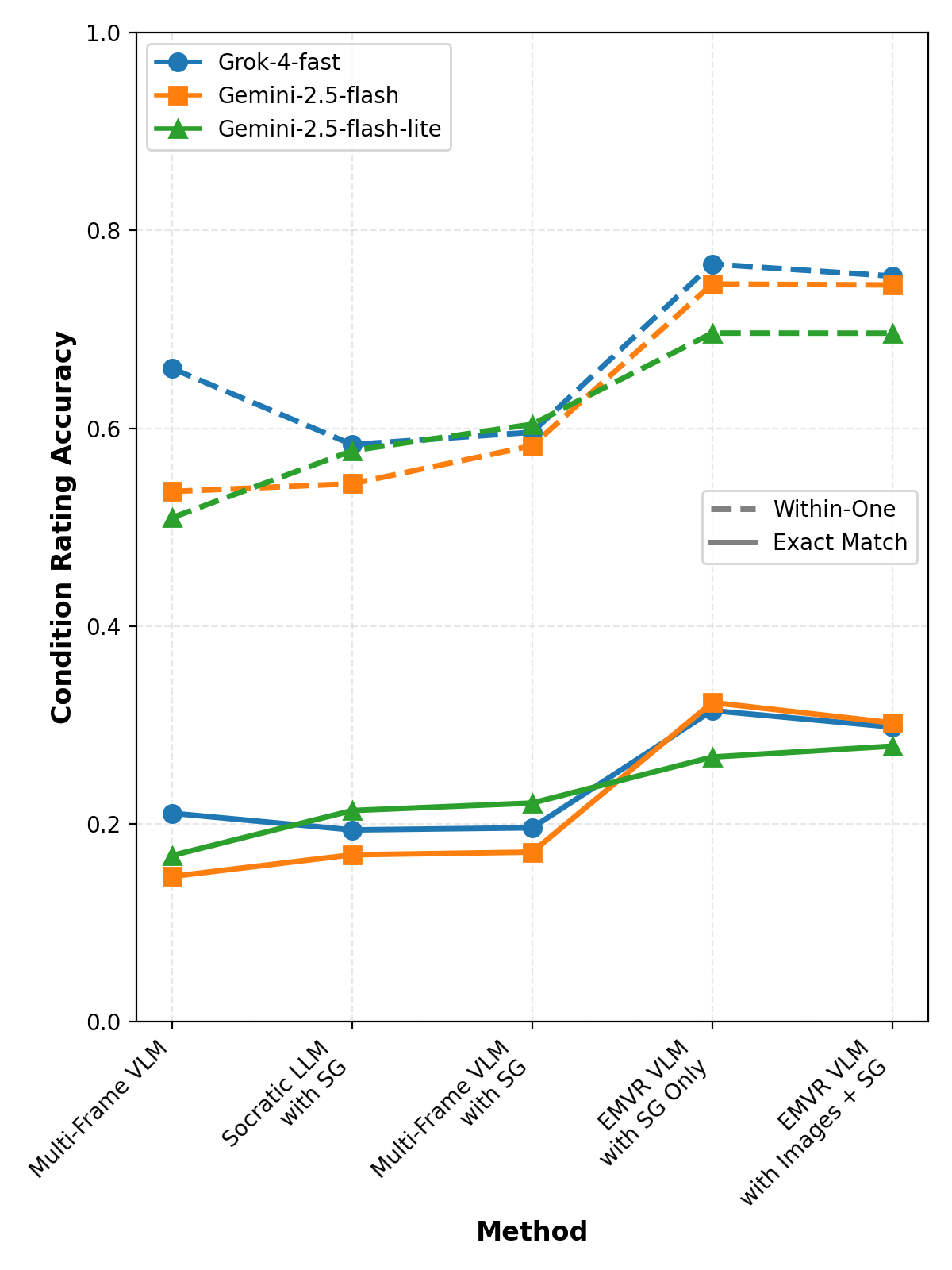}
    \caption{Condition rating prediction accuracy comparison across varying models and methods. Expert inspectors demonstrate \humanUpperLimitConditionRatingBaseline{} consistency between assigned ratings when ratings fall within $\pm 1$ of a median rating \cite{agrawal2013consistency,bridge_inspection_accuracy}.}
    \label{fig:condition_rating_accuracy}
\end{figure}

Figure~\ref{fig:condition_rating_accuracy} presents the condition rating prediction accuracy across all three VLMs and five methods. The results demonstrate that EMVR improves performance across multiple metrics and models.

Tables~\ref{tab:answer_correctness} and \ref{tab:image_selection} present comprehensive performance metrics across all configurations and models. The EMVR VLM w/ Images + SG configuration achieves strong performance across both metrics. For Answer Correctness, Grok 4 Fast reaches 0.648 while Gemini 2.5 Flash EMVR VLM w/ SG Only achieves 0.609. Particularly notable is the visual grounding performance: Grok 4 Fast EMVR VLM w/ Images + SG achieves 0.889 \imageMetricName{}, demonstrating strong capability in identifying relevant visual evidence.

\begin{table}[t]
\centering
\caption{Answer Correctness across three VLMs and five methods.}
\label{tab:answer_correctness}
\scriptsize
\begin{tabular}{p{3cm}ccc}
\hline
 \textbf{Method} & \textbf{Gemini 2.5} & \textbf{Gemini 2.5} & \textbf{Grok 4 Fast} \\
  & \textbf{Flash Lite} & \textbf{Flash} &  \\
\hline
 Multi-Frame VLM \cite{openeqa} & 0.507 & 0.484 & 0.576 \\
 Socratic LLM w/ SG \cite{openeqa,zeng2022socratic} & 0.542 & 0.588 & 0.614 \\
 Multi-Frame VLM w/ SG \cite{openeqa}& \textbf{0.581} & 0.548 & 0.622 \\
 EMVR VLM w/ SG Only & 0.512 & \textbf{0.609} & 0.638 \\
 EMVR VLM w/ Images + SG & 0.497 & 0.551 & \textbf{0.648} \\
\hline
\end{tabular}
\end{table}

\begin{table}[t]
\centering
\caption{\imageMetricName{} across three VLMs and five methods.}
\label{tab:image_selection}
\scriptsize
\begin{tabular}{p{3cm}ccc}
\hline
 \textbf{Method} & \textbf{Gemini 2.5} & \textbf{Gemini 2.5} & \textbf{Grok 4 Fast} \\
  & \textbf{Flash Lite} & \textbf{Flash} &  \\
\hline
 Multi-Frame VLM \cite{openeqa} & 0.717 & 0.694 & 0.687 \\
 Socratic LLM w/ SG \cite{openeqa,zeng2022socratic} & 0.775 & 0.767 & 0.838 \\
 Multi-Frame VLM w/ SG \cite{openeqa}& 0.802 & 0.778 & 0.833 \\
 EMVR VLM w/ SG Only & 0.798 & \textbf{0.836} & 0.876 \\
 EMVR VLM w/ Images + SG & \textbf{0.849} & 0.803 & \textbf{0.889} \\
\hline
\end{tabular}
\end{table}

\subsection{Error Analysis and Failure Modes}\label{subsec:error_analysis}

We perform a qualitative analysis, presented in Figure~\ref{fig:qualitative}, where we showcase a successful example and contrast it with two commonly observed failure cases. In the successful example, the agent identifies the correct substructure and uses relevant image citations to ground its answer, achieving a condition rating within $\pm 1$ of the ground truth. Conversely, we identified two primary failure modes that account for the large majority of incorrect condition assessments. The first is poor image citations, where the agent cites irrelevant images, leading to an incorrect answer and rating. The second is hallucinated image citations, where a VLM invents citations for images that do not exist, resulting in nonsensical generations. These findings indicate that low quality image citations can be a proxy for detecting the occurrence of hallucinations or poor answer generations.

\subsection{Limitations}
While we tested several sub-30B-parameter open-source VLMs, they could not reliably adhere to the required structured-output and function-calling formats. These models also have substantially lower context windows, making evaluation on larger scenes infeasible. We therefore excluded them from the main comparison to avoid unfair evaluations, but we provide results where applicable in the Supplementary Material.



\section{Conclusion}
\label{sec:conclusion}

In this work, we introduced \textbf{BridgeEQA}, a real-world Embodied Question Answering benchmark grounded in professional bridge inspection, comprising 2,200 question-answer pairs across 200 bridge scenes with 9,586 images. By leveraging egocentric imagery, expert-authored reports, and standardized NBI condition ratings, the dataset provides a testbed for evaluating spatial reasoning and multi-scale evidence aggregation in a domain with measurable expert-level criteria. To assess visual grounding, we proposed \textbf{\imageMetricName{}}, a metric that measures semantic alignment between agent-cited images and reference evidence sets. We further presented \textbf{EMVR}, an EQA method that reformulates Episodic Memory EQA as traversal over an image-based scene graph, enabling dynamic context retrieval rather than fixed long-context input. Evaluations show improvements with EMVR across metrics. Using Grok 4 Fast, we find that EMVR improves condition rating accuracy within~$\pm 1$ by 9.3 percentage point, \imageMetricName{} by 20.2 percentage point, and Answer Correctness by 7.2 percentage point over the Multi-Frame VLM baseline.

\section{Acknowledgment}
Authors acknowledge partial financial support from the Texas Department of Transportation grant number 0–7181.
\newpage
{
    \small
    \bibliographystyle{ieeenat_fullname}
    \bibliography{main}
}
\appendix
\setcounter{page}{1}
\setcounter{section}{0}
\maketitlesupplementary



\section{Evaluating \imageMetricName{} for Human Alignment}
\label{sec:rationale}

To validate that our \imageMetricName{} metric aligns with human judgment, we conducted a manual annotation study on a randomly sampled set of 100 question-answer pairs from BridgeEQA. For each sample, we randomly perturbed the reference images by introducing varying numbers of random images from the original PDF document set and then randomly removing a varying number of images. This perturbation process generated image sets spanning the full relevance spectrum—from completely irrelevant to fully relevant to the question.

Three annotators independently labeled each sample on a 5-point scale, which we normalized to the 0.0-1.0 range to match the \imageMetricName{} output range. We then computed \imageMetricName{} scores for the same dataset using Gemini-2.5-flash as the evaluator model.

The Spearman correlation between the averaged human annotations and \imageMetricName{} scores was 0.817, demonstrating strong alignment between our automated metric and human judgment of image relevance.

\section{Dataset Example}
\label{sec:dataset_examples}

We provide an example from BridgeEQA in Figure \ref{fig:bridge_eqa_example}. The reference images are images parsed from the source report and the condition rating is extracted from the answer. This particular scene graph for structure 0010 in Chelsea has a total of 53 nodes.

\begin{figure*}[!t]
    \centering
    
    \begingroup
    \setlength{\parskip}{0pt}
    \begin{tcolorbox}[
        colback=gray!5!white, 
        colframe=gray!60!black,
        title={BridgeEQA Example: Structure 0010 in Chelsea},
        width=\textwidth
    ]
    
    \small
    
    \textbf{Question:}  
    Given the transverse cracking in the asphalt wearing surface where no joints are present, what are the resulting deterioration patterns—such as cracking and abrasion—observed on the abutments, and what is the overall condition rating for the substructure?
    
    \vspace{0.5em}
    
    \textbf{Answer:}  
    Asphalt over the structure has transverse cracking forming in various areas scattered throughout both sides over the abutments. The substructure below shows related deterioration. Reinforced concrete abutment \#1 has multiple settlement cracks present, and several areas along the lower portions exhibit minor abrasion. Reinforced concrete abutment \#2 has a full-height vertical crack, and its lower portions have heavy abrasion/scaling that penetrates as deep as approximately 15". Because all primary structural elements are sound but have minor section loss and cracking, the substructure is in \textbf{Fair Condition} with a rating of \textbf{5}.
    
    \vspace{0.75em}
    
    \textbf{Reference Images:}
    \begin{itemize}[leftmargin=2em, topsep=0pt]
        \item 636045817baf4010d986b01a55baca1d.png
        \item 1de1058cd997112823da497df50cc66b.png
        \item 11a2bf060c346794067d43bc7718caad.png
        \item 3c435b2e6814c354430fee1480befc8d.png
    \end{itemize}
    
    \vspace{0.25em}
    
    \textbf{Extracted Condition Rating:} 5
    
    \vspace{0.75em}
    
    \textbf{Scene Graph:}
    \begin{verbatim}
    {
      "nodes": [
        {
          "image_name": "13b13f0ccab54674d56471f83082c47c.png",
          "central_focus": "Upstream view of Jail Brook from under the bridge.",
          "image_description": "View from under the bridge looking upstream...",
          "edges": [
            {
              "connected_to": "95b6875b4d3e1bcc4d524926a13b0157.png",
              "description_of_connection": "adjacent_to"
            },
            {
              "connected_to": "be36bcfe248d002bf1b6d2a62e6b20ae.png",
              "description_of_connection": "provides_context_for"
            }
          ]
        },
        ...
      ]
    }
    \end{verbatim}
    
    \end{tcolorbox}
    \endgroup
    \caption{We provide an example from BridgeEQA on structure 0010 in Chelsea. This particular example has a scene graph with 53 nodes.}
    \label{fig:bridge_eqa_example}

\end{figure*}

\section{Dataset Creation Details}
\label{sec:dataset_examples}

\subsection{Example Source Inspection Reports}
\label{subsec:example_inspection_reports}

\begin{figure*}
  \centering
  \includegraphics[width=\textwidth]{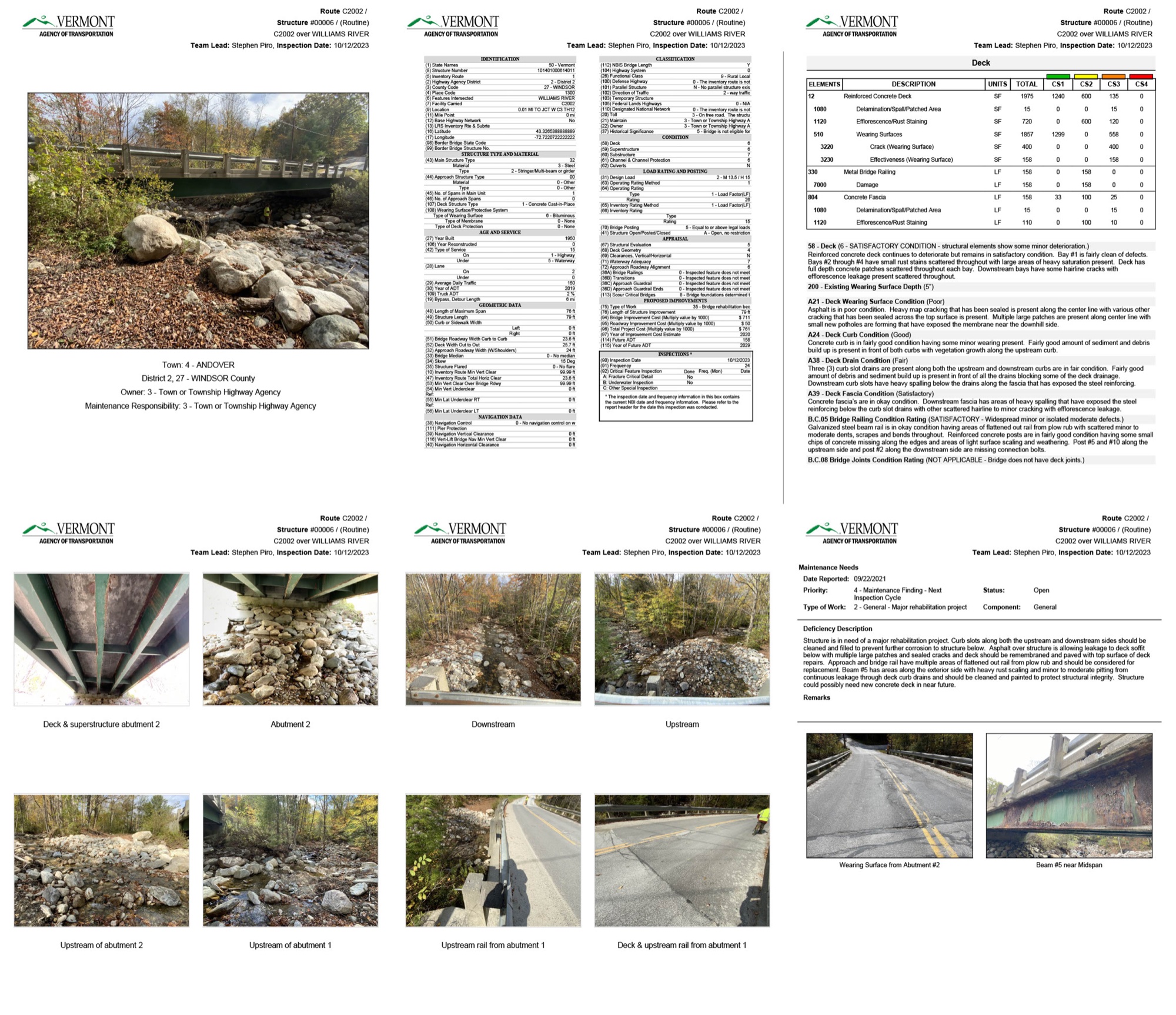}
  \caption{Sample pages of BridgeEQA source report originating from the Vermont Agency of Transportation's (VTrans) inspection report for Structure 00006, located in Andover.}
  \label{fig:vermont_report_example}
\end{figure*}

We provide sample pages from Vermont bridge inspection report in Figure \ref{fig:vermont_report_example} as an example source report for BridgeEQA.

\subsection{Data Collection}

We collected bridge inspection reports from the Vermont Agency of Transportation (VTrans) public database, which contains unstructured PDF inspection reports covering bridges across Vermont. Each report documents the condition of a single unique bridge and includes inspector observations, condition ratings, and photographic documentation.

\subsection{Stage 1: Preprocess and Filter}

The preprocessing stage applies quality control filters to ensure that selected reports contain sufficient visual documentation for a meaningful infrastructure assessment. 

\textbf{Report-Level Filtering.} Initial qualitative evaluation of the inspection reports revealed significant variability in visual documentation quality and comprehensiveness. To ensure sufficient visual coverage for meaningful condition assessment, we applied a minimum threshold of 20 images per report. Reports failing to meet this criterion were excluded based on several quality indicators:

\begin{itemize}
    \item \textbf{Incomplete visual coverage:} Reports with fewer images often documented only limited perspectives of the bridge, missing critical structural components necessary for comprehensive assessment.
    \item \textbf{Low image quality:} Sparse image sets frequently exhibited poor resolution, unfavorable lighting conditions, or obstructed views that would hinder reliable condition evaluation.
    \item \textbf{Non-standard outlier conditions:} Some reports documented bridges that were demolished, under major reconstruction, or otherwise not representative of typical operational infrastructure.
\end{itemize}

\textbf{Page Filtering.} Page filtering removes the first two pages of each report, which typically contain administrative cover pages, title pages, and summary information without detailed inspection content or photographic documentation.

\textbf{Image-Level Filtering.} Within the quality-controlled reports, individual images underwent additional filtering. Images smaller than 200x200 pixels were systematically removed from the dataset. This threshold was established through empirical observation that sub-threshold images predominantly contained:

\begin{itemize}
    \item Organizational logos and branding elements
    \item Document headers and administrative markings
    \item Thumbnails and preview images lacking structural detail
\end{itemize}

Such images provide minimal information for infrastructure condition assessment and could introduce noise into model training or evaluation.

\textbf{Random Sampling.} From the filtered pool of quality-controlled reports, we employed a random sampling strategy to select \numReportsFiltered{} reports for the final dataset. This sampling approach ensures representative coverage of Vermont's bridge inventory while maintaining computational tractability for annotation and evaluation.

\subsection{Stage 2: Extract}

The extraction stage processes filtered PDFs to obtain textual and visual content. Text extraction parses inspector notes, observations, and structured fields. Image extraction retrieves photographs meeting quality criteria, preserving metadata about location and context. This stage yielded \numImages{} images across \numReportsFiltered{} reports, averaging \avgImagesPerReport{} per report.

\subsection{Stage 3: Transform}

The transformation stage structures the extracted content into standardized formats with ground truth annotations. We employ several vision-language models as zero-shot parsing tools to extract structured information from the inspection reports. Gemini 2.5 Flash and Gemini 2.5 Pro~\cite{comanici2025gemini} serves as the primary extraction model for its efficiency and quality, with no fine-tuning or training on the bridge inspection data. The models function purely as information extraction tools, parsing existing content rather than learning dataset-specific patterns. We found Gemini 2.5 Flash to frequently have parsing errors or hallucinations as context size's grew in this stage, as such we fall back to Gemini 2.5 pro to reprocess when these errors occur and drop reports if errors persist.

\textbf{Image Reference Mapping.} This component links photographs to corresponding textual descriptions in inspector notes, supporting scene formation where multiple images document the same infrastructure component. This step is required to allow grounded questions that use real references for component names, such as Abutment 1.

\textbf{Condition Rating Extraction.} This component parses component-level NBI ratings from inspector assessments, providing ground truth labels on the standardized 0-9 scale~\cite{fhwa1995coding}.

\textbf{Inspector Note Preservation.} Inspector notes are preserved to maintain the original context and rationale for condition assessments, ensuring that ground truth annotations remain grounded in the source documentation. We leverage these notes to ensure QA generation is grounded to real statements in the report.

\subsection{Stage 4: Validate}

Human quality control checks verify data integrity before QA generation. Additionally we test for any false or hallucinated image references. Parsing error detection identifies reports with corrupted text extraction, malformed condition ratings, or broken image-text mappings. When parsing errors or missing image references are detected, the report is automatically reprocessed using Gemini 2.5 Pro~\cite{comanici2025gemini} as a fallback model for more robust extraction. Both Flash and Pro are used solely as zero-shot parsing tools without any training or fine-tuning, ensuring that evaluation results reflect genuine visual reasoning capabilities rather than memorization. Reports that fail validation after reprocessing were removed from the dataset.

\subsection{Stage 5: Generate QA}

The final stage generates structured question-answer pairs for evaluation. Using a Gemini 2.5 Flash and Pro, we create questions grounded in the inspection report content, spanning condition assessment, component identification, and defect description tasks. Each answer includes the ground truth response sourced from inspector notes, references to supporting images, and the associated NBI condition rating when applicable. Quality checks verify that all referenced images exist and that answers are properly grounded in the available evidence.

\subsection{Data Generation Validation}

To ensure QA quality, we employ several evaluation metrics. We use the RAGAs \cite{es-etal-2024-ragas} metrics: Faithfulness, which measures how well answers are grounded in the provided context, and Answer Relevancy, which assesses how effectively answers address the posed questions. We also incorporate the Answerability metric from RAGalyst \cite{gao2025ragalyst} to determine whether questions can be adequately answered given the available context. To assess domain specificity, we employ LLM-as-a-Judge to determine Inspector Relevancy (0.0-1.0). This score measures the direct applicability of the question and its associated answer for bridge inspectors.

After evaluating all QA's with Gemini-2.5-flash, we reach a Faithfulness of $0.997$, an Answer Relevancy of $0.997$, an Answerability of $0.996$, and an Inspector Relevancy of $0.980$. These high scores across all metrics demonstrate the overall high quality of the dataset.

\subsection{Human Validation}

To validate the automated filtering and processing pipeline, human evaluation was conducted on a random subsets of the processed reports. This manual inspection verified that the quality-controlled reports met the following standards:

\begin{itemize}
    \item Sufficient visual coverage of critical bridge components
    \item Adequate image quality for condition assessment
    \item Consistency with typical operational bridge inspection documentation
    \item Accurate representation of the condition rating labels
\end{itemize}

\section{Effects of Scene Graph Connectivity on Condition Rating}
\label{sec:scene_graph_details}
\begin{figure*}
  \centering
  \includegraphics[trim={0mm 0mm 0mm 10mm}, clip, width=\textwidth]{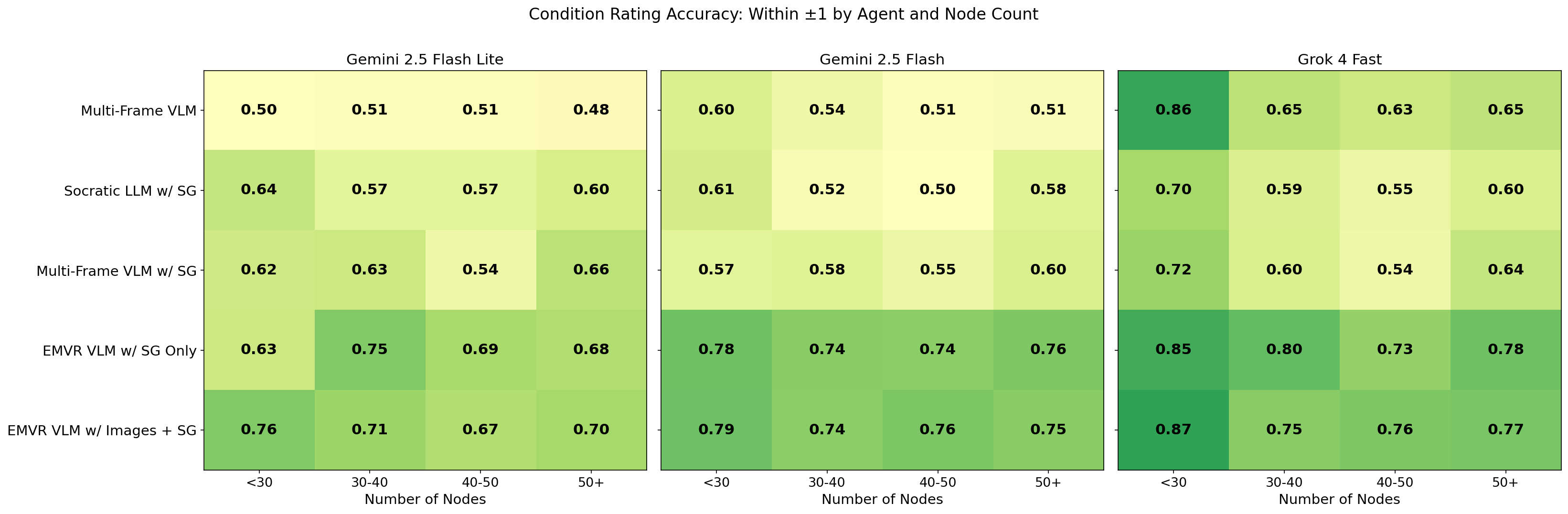}
  \caption{Condition rating within $\pm$ 1 accuracy heat map across method, VLM, and number of nodes.}
\label{fig:heatmap_nodes}
\end{figure*}

\begin{figure*}
  \centering
  \includegraphics[trim={0mm 0mm 0mm 10mm}, clip, width=\textwidth]{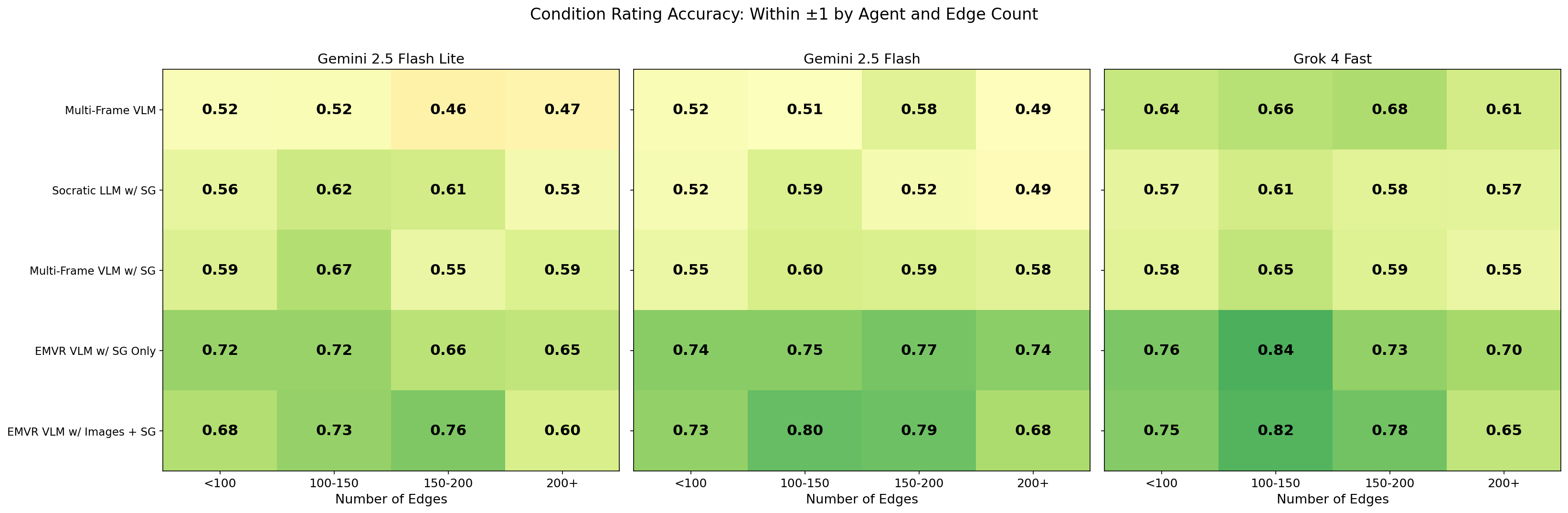}
  \caption{Condition rating within $\pm$ 1 accuracy heat map across method, VLM, and number of edges.}
\label{fig:heatmap_edges}
\end{figure*}

We provide the accuracy heatmap of each method and VLM by the number of nodes in the scene graph in Figure~\ref{fig:heatmap_nodes} and by the number of edges in the scene graph in Figure~\ref{fig:heatmap_edges}.

Generally, the performance across methods decreases as the number of edges and nodes increase. This is due to the increased context sizes which is known to reduce VLM performance. However, EMVR performance degrades less at higher node and edge counts since EMVR mitigates the "lost in the middle" problem as explained in Figure \ref{fig:context_lost_in_the_middle}.  

\section{Open-Source Model Results}
\label{sec:open_source_results}

We extend our evaluation to open-source VLMs (Vision-Language Models) to assess generalizability beyond proprietary models. Given the large context windows required by our dataset, we omit Multi-Frame VLM w/ SG and EMVR VLM w/ Images + SG, as both require encoding images alongside the scene graph, which exceeds the context window of these models. Additionally, these models exhibited high failure rates in structured output generation, hallucinated function calls, and repeated the same actions in loops during agent execution. Due to these limitations, only a fraction of BridgeEQA could be tested. Results in \ref{tab:exact_match} and \ref{tab:within_one} should therefore not be compared against the main paper results.

\begin{table}[t]
\centering
\caption{Condition rating exact match accuracy (\%) on open-source VLMs evaluated on BridgeEQA instances with fewer than 30 images.}
\label{tab:exact_match}
\scriptsize
\begin{tabular}{p{2cm}ccc}
\hline
 & \textbf{Qwen3-VL} & \textbf{Qwen3-VL} & \textbf{Nemotron-3} \\
 \textbf{Method} & \textbf{8B-Thinking}\cite{qwen3vl}$^\dagger$ & \textbf{30B-A3B}\cite{qwen3vl} & \textbf{Nano-30B-A3B}\cite{nemotron3} \\
\hline
Multi-Frame\newline VLM & 9.1 & 8.5 & 11.0 \\
Socratic LLM\newline w/ SG & 36.4 & 6.1 & 11.0 \\
EMVR VLM\newline w/ SG Only & 40.9 & \textbf{29.3} & \textbf{23.2} \\
\hline
\multicolumn{4}{p{0.95\linewidth}}{\footnotesize $^\dagger$Due to its limited context window, the Qwen3-VL 8B model was evaluated only on scenes with fewer than 30 images, a small fraction of the full dataset. These results are not directly comparable to other models reported in this paper.}
\end{tabular}
\end{table}
\begin{table}[t]
\centering
\caption{Condition rating within $\pm$1 accuracy (\%) on open-source VLMs evaluated on BridgeEQA instances with fewer than 30 images.}
\label{tab:within_one}
\scriptsize
\begin{tabular}{p{2cm}ccc}
\hline
 & \textbf{Qwen3-VL} & \textbf{Qwen3-VL} & \textbf{Nemotron-3} \\
 \textbf{Method} & \textbf{8B-Thinking}\cite{qwen3vl}$^\dagger$ & \textbf{30B-A3B}\cite{qwen3vl} & \textbf{Nano-30B-A3B}\cite{nemotron3} \\
\hline
Multi-Frame\newline VLM & 81.8 & 23.2 & 58.5 \\
Socratic LLM\newline w/ SG & 72.7 & 30.5 & 48.8 \\
EMVR VLM\newline w/ SG Only & 81.8 & \textbf{76.8} & \textbf{70.7} \\
\hline
\multicolumn{4}{p{0.95\linewidth}}{\footnotesize $^\dagger$Due to its limited context window, the Qwen3-VL 8B model was evaluated only on scenes with fewer than 30 images, a small fraction of the full dataset. These results are not directly comparable to other models reported in this paper.}
\end{tabular}
\end{table}

\end{document}